\documentclass[letterpaper]{article} 
\usepackage[submission]{aaai25}  
\usepackage{amsmath,amssymb}
\usepackage{times}  
\usepackage{helvet}  
\usepackage{courier}  
\usepackage[hyphens]{url}  
\usepackage{graphicx} 
\usepackage{natbib}  
\usepackage{caption} 
\usepackage{algorithm}
\usepackage{algorithmic}

\usepackage{subfigure}
\usepackage{todonotes}
\usepackage{multirow}
\usepackage{xcolor}
\usepackage{color}
\usepackage{tabularx}
\usepackage{textcomp}
\usepackage{booktabs}
\usepackage{makecell}
\usepackage{pifont}

\DeclareMathOperator*{\argmin}{arg\,min}

\usepackage{newfloat}
\usepackage{listings}
\DeclareCaptionStyle{ruled}{labelfont=normalfont,labelsep=colon,strut=off} 
\floatstyle{ruled}
\newfloat{listing}{tb}{lst}{}
\floatname{listing}{Listing}
\title{E4: Energy-Efficient DNN Inference for Edge Video Analytics \\ Via Early-Exit and DVFS}
\author {
    Ziyang Zhang\textsuperscript{\rm 1},
    Yang Zhao\textsuperscript{\rm 2}\thanks{Corresponding Author},
    Ming-Ching Chang\textsuperscript{\rm 3},
    Changyao Lin\textsuperscript{\rm 1},
    Jie Liu\textsuperscript{\rm 1,\rm 2}
}
\affiliations {
    \textsuperscript{\rm 1}UbiCIS, Faculty of Computing, Harbin Institute of Technology\\
    \textsuperscript{\rm 2}International Research Institute for Artificial Intelligence, Harbin Institute of Technology (Shenzhen)\\
    \textsuperscript{\rm 3}Department of Computer Science, University at Albany - SUNY\\
    zhangzy@stu.hit.edu.cn, yang.zhao@hit.edu.cn, mchang2@albany.edu, lincy@stu.hit.edu.cn, jieliu@hit.edu.cn
}

\usepackage{bibentry}

\begin{document}

\maketitle

\begin{abstract}
Deep neural network (DNN) models are increasingly popular in edge video analytic applications. 
However, the compute-intensive nature of DNN models pose challenges for energy-efficient inference on resource-constrained edge devices.
Most existing solutions focus on optimizing DNN inference latency and accuracy, often overlooking energy efficiency.
They also fail to account for the varying complexity of video frames, leading to sub-optimal performance in edge video analytics.
In this paper, we propose an Energy-Efficient Early-Exit (\texttt{E4}) framework that enhances DNN inference efficiency for edge video analytics by integrating a novel early-exit mechanism with dynamic voltage and frequency scaling (DVFS) governors. 
It employs an attention-based cascade module to analyze video frame diversity and automatically determine optimal DNN exit points. 
Additionally, \texttt{E4} features a just-in-time (JIT) profiler that uses coordinate descent search to co-optimize CPU and GPU clock frequencies for each layer before the DNN exit points. Extensive evaluations demonstrate that \texttt{E4} outperforms current state-of-the-art methods, achieving up to 2.8$\times$ speedup and 26\% average energy saving while maintaining high accuracy.
\end{abstract}

\section{Introduction}
\label{intro}

Advances in deep neural network (DNN) models and GPU hardware accelerators have significantly advanced video analytics in edge intelligence applications, including object detection~\cite{zou2023object,zhao2019object}, action recognition~\cite{ghodrati2021frameexit,jhuang2013towards}, and pose estimation~\cite{andriluka20142d,toshev2014deeppose,andriluka2018posetrack}, {\em etc.}
To protect data privacy and ensure low-latency quality of service (QoS), many of these applications are deployed on edge devices close to the data sources~\cite{liang2023dvelen}. However, the increasing demand for higher video quality results in greater video frame complexity, making DNN models computationally intensive for tasks like multi-object detection and tracking.
On the hardware side, edge devices face limitations in cost and size, resulting in fewer computational resources compared to cloud servers~\cite{bhardwaj2022ekya,padmanabhan2023gemel,khani2023recl}.
The diverse network structures of DNN models~\cite{cui2022dvabatch} and the varying complexity of video frames~\cite{menon2022vca} ({\em e.g.}, the spatial correlation between consecutive frames) introduce new challenges for edge video analytics.

Dynamic voltage and frequency scaling (DVFS) is a widely used power management technique that balances energy consumption and computing performance on edge devices by adjusting CPU and GPU voltage-frequency in real-time. While previous studies have developed various learning-based DVFS governors~\cite{kim2021ztt,yeganeh2020ring,lin2023workload}, we find that their performance suffers when applied to edge video analytics due to a mismatch between CPU/GPU frequency settings and the specific demands of video processing at the edge.

\begin{figure}[t]
\centerline{
\subfigure[Inference latency (ms)]{
\begin{minipage}[b]{0.49\linewidth}
\includegraphics[width=1\linewidth]{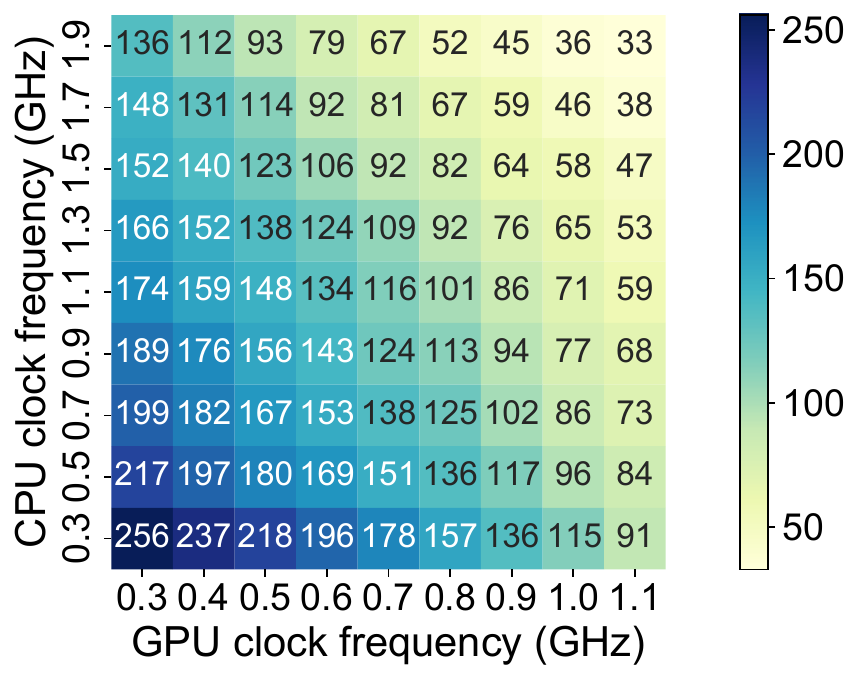}
\end{minipage}}
\subfigure[Energy consumption (W)]{
\begin{minipage}[b]{0.47\linewidth}
\includegraphics[width=1\linewidth]{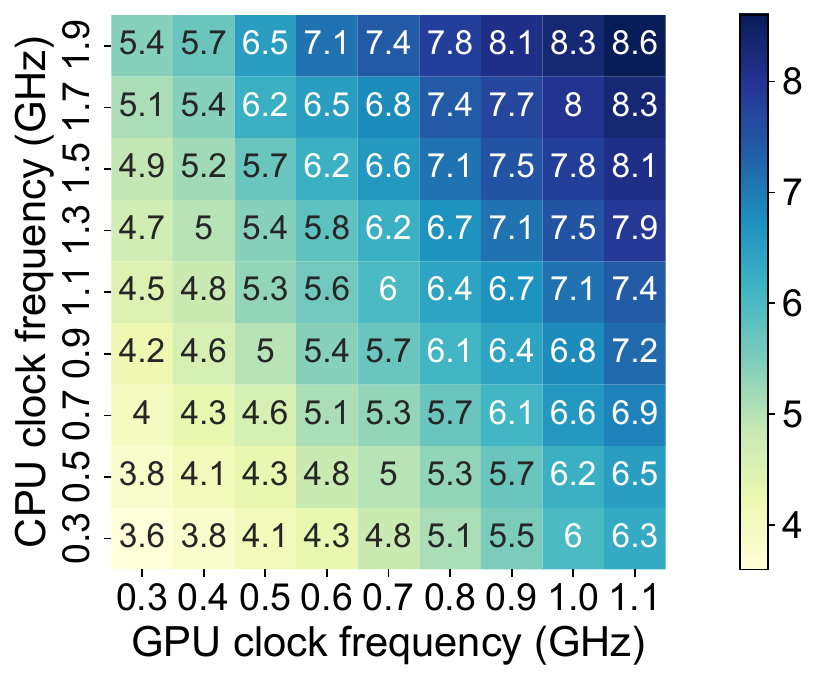}
\end{minipage}}
}
\caption{The impact of CPU and GPU clock frequencies on (a) inference latency (ms) and (b) energy consumption (W), based on running EfficientNet-B0 on an Nvidia Xavier NX edge GPU with 8GB DRAM.
}
\label{observation}
\end{figure}

As a motivating example, we tested \emph{zTT}~\cite{kim2021ztt}, a state-of-the-art learning-based DVFS governor, on an Nvidia Jetson Xavier NX edge device running the EfficientNet-B0 DNN model~\cite{tan2019efficientnet}. The results, shown in Fig.~\ref{observation}, highlight the trade-off between inference latency and energy consumption. Specifically, we observe that {\em higher processor clock frequencies reduce inference latency but significantly increase energy consumption.}
For instance, achieving real-time video analytics at 30 frames per second (fps), which requires a 30ms inference latency, necessitates setting the CPU and GPU frequencies to their highest levels of 1.9GHz and 1.1GHz, respectively. However, this leads to a sharp rise in energy consumption of 8.6W. Similar trends were observed on other edge devices as well.

\begin{figure*}[t]
\centering
\centerline{\includegraphics[width=0.8\linewidth]{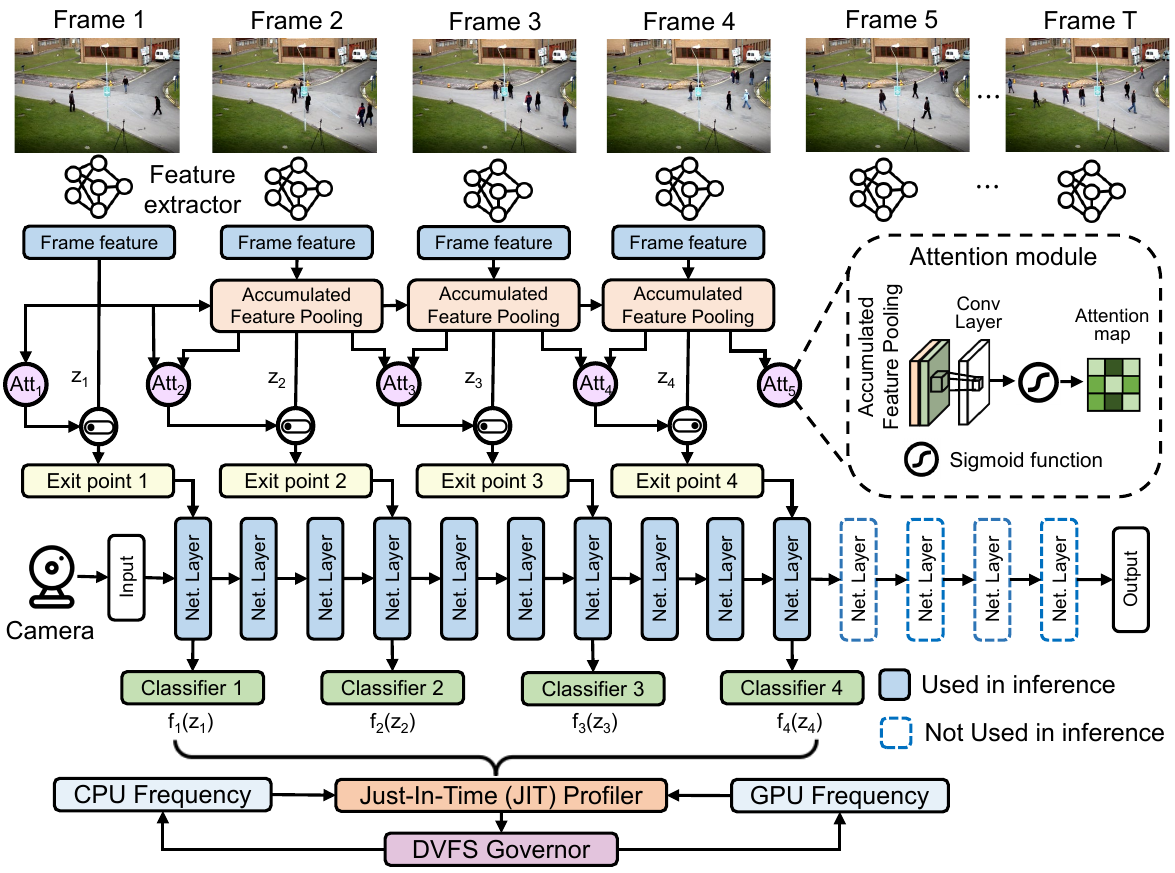}}
\caption{Overview of the proposed \texttt{E4} efficient edge DNN video analytic inference framework. Given a video input, we sample $T$ frames with varying complexities, such as different numbers of detectable objects. The feature extractor processes each frame and aggregates these features to assess video frame complexity. An attention module and its corresponding gate are trained to determine DNN early exit points. The Just-In-Time (JIT) Profiler and DVFS Governor are then employed to search and scale CPU and GPU clock frequencies for each layer before the DNN exit points. 
}
\label{fig:system}
\end{figure*}

Based on the experiments and observations, we identified that the challenges stem from the diversity in video frame complexity and DNN models. As illustrated in Fig.~\ref{fig:system}, different video frames vary in the number of objects they contain, leading to varying levels of complexity for object detection. We use the term {\em video frame complexity} to describe these differing DNN inference demands in general video analytics applications.
State-of-the-art DNN models are designed to detect multiple objects with high accuracy, but not all network layers are necessary for frames with fewer objects. For these low-complexity frames, DVFS should adaptively reduce CPU and GPU clock frequencies. However, current learning-based DVFS approaches do not account for video frame complexity.
Additionally, edge video analytics often deploy different DNN models for various tasks like detection and tracking, which further complicates performance optimization. Conventional DVFS methods are application-agnostic and do not consider the specific workload characteristics of DNNs. In this paper, our proposed framework addresses both video frame complexity and DNN model diversity to optimize edge performance.

In this paper, we introduce the {\bf Energy-Efficient Early-Exit (\texttt{E4})} framework that improves DNN inference efficiency and latency for edge video analytics. Early-Exit is a mechanism that adaptively exits DNN inference early based on video frame complexity and DNN model diversity~\cite{teerapittayanon2016branchynet,laskaridis2020spinn,zhang2023octopus}. 
We introduce two key design implementing \texttt{E4}:
(1) an attention-based cascade module that determines DNN exit points by analyzing video frame complexity, and (2) a novel DVFS governor that automatically adjusts CPU and GPU frequencies for each layer before the DNN exit points using a Just-In-Time (JIT) profiler~\cite{you2023zeus} based on coordinate descent search.
We conduct comprehensive experiments on two widely-used datasets and two representative video analytics DNN models across five heterogeneous edge devices.
The evaluation results demonstrate that \texttt{E4} achieves lower latency and higher energy efficiency compared to state-of-the-art methods, while maintaining accuracy.
The main contributions of this paper are as follows:
\begin{itemize}
\item We propose \texttt{E4}, an energy-efficient inference framework that enables adaptive DNN exit points and power management configurations tailored to dynamic video frames and DNN models.

\item We design an attention-based cascade module that determines optimal exit points by analyzing the spatial correlation between consecutive video frames.

\item We develop a novel power management approach using the coordinate descent search algorithm to automatically scale CPU and GPU frequencies for each network layer. Additionally, a lightweight DNN-based prediction model minimizes performance interference between multiple DNN models.

\item Extensive experimental results demonstrate that \texttt{E4} outperforms state-of-the-art early-exit methods, achieving up to 2.8× speedup and 26\% average energy savings.

\end{itemize}

\section{Related Works}
\label{Related}

\noindent
\textbf{Power management on edge devices:} 
Previous works have introduced various DVFS governors for power management on edge devices. For instance, zTT~\cite{kim2021ztt}, GearDVFS~\cite{lin2023workload} 
and Ring-DVFS~\cite{yeganeh2020ring} use learning-based approaches to automatically optimize CPU and GPU frequencies to reduce energy consumption, often at the cost of inference performance.
In contrast, methods like Road-RuNNer~\cite{kakolyris2023road}, DVFO~\cite{zhang2024dvfo,zhang2023dvfo} and AppealNet~\cite{li2021appealnet} use adaptive partitioning to enable cloud-edge collaborative inference, reducing energy consumption by running parts of DNN models on edge devices. Zeus~\cite{you2023zeus} focuses on optimizing energy during DNN training with a multi-arm bandit-based power optimizer that balances energy consumption and latency. Additionally, techniques like model compression~\cite{han2015deep} and neural architecture search (NAS)~\cite{ren2021comprehensive} are complementary to \texttt{E4} and can further enhance energy efficiency by using lightweight DNN models.

\medskip
\noindent
\textbf{Early-exit DNN inference:} 
The early-exit mechanism~\cite{teerapittayanon2016branchynet} is a type of dynamic inference that allows DNNs to exit at different layers or sub-networks, once an accuracy threshold is met, thus reducing computation costs while maintaining accuracy. 
Since its introduction in BranchNet~\cite{teerapittayanon2016branchynet}, various early-exit strategies have been developed. 
For instance, HarvNet~\cite{jeon2023harvnet} uses NAS to automatically determine exit points, Delen~\cite{liang2023dvelen} employs conditional execution for adaptive control of latency, accuracy, and power. PAME~\cite{zhang2022pame} focuses on reducing batch inference latency with precision-aware early exits.
However, these approaches do not integrate power management techniques like DVFS for energy-efficient dynamic inference. While methods such as EdgeBERT~\cite{tambe2021edgebert}, EENet~\cite{li2023eenet} and Predictive Exit~\cite{li2023predictive} combine early exits with DVFS, they overlook the impact of video frame complexity on DNN inference. In contrast, our work analyzes video frame complexity to optimize energy consumption and inference latency, while maintaining DNN inference accuracy in edge video analytics.

\medskip
\noindent
\textbf{Edge video analytics:} 
DNN-based video analysis on edge devices near the data source is a promising approach. 
Previous works have focused on optimizing inference latency, accuracy, and memory overhead. For example, Ekya~\cite{bhardwaj2022ekya}, RECL~\cite{khani2023recl} and AdaInf~\cite{shubha2023adainf} use continuous learning~\cite{wang2024comprehensive} for incremental training on edge devices, which addresses accuracy degradation from data drift.
Gemel~\cite{padmanabhan2023gemel} reduces memory overhead through model merging.
Remix~\cite{10.1145/3447993.3483274} partitions video frames by the number of objects, using complex DNN models for regions with more objects and simpler models for regions with fewer objects to reduce inference latency.
These approaches are complementary to 
\texttt{E4}, which further enhances inference performance.

\section{Methodology}

\subsection{Design Overview}

Our proposed \texttt{\texttt{E4}} framework addresses energy-efficient DNN inference for edge video analytics. The core idea of \texttt{E4} is to determine optimal exit points for different video frames based on their complexity. A learning-based DVFS governor then co-optimizes CPU and GPU clock frequencies for each network layer before these exit points during the DNN inference. 
As illustrated in Fig.~\ref{fig:system}, \texttt{E4} consists of four key components: (1) \emph{Accumulated Feature Pooling} analyzes video frame complexity. (2) \emph{Attention-Based Early-Exit} determines the appropriate DNN exit points. (3) {\em Just-In-Time (JIT) Profiler} dynamically co-optimizes CPU and GPU frequencies for each layer before the exit points. (4) \emph{DVFS governor} manages dynamic voltage and frequency scaling. The framework uses historical video frame data to train the early-exit DNN model, leveraging the temporal correlation between frames. Each classifier performs inference at the layer corresponding to each early exit point.
This dynamic power management approach makes informed decisions based on the complexity of video frames and DNN models. Details of each component are discussed in the following sections.

\subsection{Accumulated Feature Pooling}

Given a set of video frames and their corresponding labels, $e.g.$, objects to be detected, the feature extractor generates features for each frame. In our experiments, we use the EfficientNet-B0~\cite{tan2019efficientnet} and MobileNet-v2~\cite{sandler2018mobilenetv2} as the feature extraction backbones. 
We aggregate these features using \emph{Accumulated Feature Pooling}, which is based on max-pooling~\cite{ghodrati2021frameexit}.
This method captures temporal relationships between $T$ consecutive video frames. Let $\Phi$ denote the feature extraction network parameterized by hyper-parameters $\theta$. We implement the temporal aggregation function $\Psi$ using a two-layer long short-term memory (LSTM). For a video frame $x_t$ and its extracted features $\Phi(x_t;\theta)$, the accumulated features $z_t$ of $x_t$ are obtained by:
\begin{equation}
z_t = \Psi \left( z_{t-1}, \Phi(x_t;\theta) \right).
\label{eq1}
\end{equation}

\subsection{Attention-Based Early-Exit}

We design an attention-based cascade module to extract spatio-temporal correlations between successive $T$ video frames for determining DNN exit points. Each attention module is implemented as a lightweight two-layer neural network with an 1$\times$1 convolution kernel, featuring 64 neurons in the first layer and 32 neurons in the second layer. 
Let $\beta$ denote the attention module with parameter $\sigma$ that takes the accumulated features $z_t$ and $z_{t-1}$ as input to generate attention weights. The aggregated features $\tau_t$ are obtained from the accumulated features $z_t$ as:
\begin{equation}
\tau_t = W_1 z_{t-1} \cdot \tanh(W_2 z_t),
\label{attention_1}
\end{equation}
where $W_1$ and $W_2$ are the weights to be optimized.
The normalized weight $\beta(\sigma_t)$ proportional to video frame complexity can then be obtained from the aggregated features $\tau_t$:
\begin{equation}
\beta(\sigma)_t = \frac{\exp(\tau_t)}{\sum_t \exp(\tau_t)}.
\label{eq2}
\end{equation}

The early-exit DNN model contains $E$ exit points, each equipped with a gate unit that enables early exiting. Each gate is represented as a binary decision function $\rho()$ parameterized by $\mu$, which determines whether a video frame has reached the required accuracy threshold for early exit during DNN inference. This way, each gate $\rho()$ is designed to evaluate the video frame complexity by processing the accumulated features $z_t$ and $z_{t-1}$ along with the attention weight $\beta(\sigma)_t$. Let $t^*$ denote the earliest frame that meets the exit condition. The early-exit function $e_{t^*}$ is formulated as:
\begin{equation}
e_{t^*} = \argmin_t \rho \left(
(z_{t-1}, z_t,\beta(\sigma)_t);\mu
\right).
\label{eq3}
\end{equation}
If no gate meets the exit condition, the entire DNN model is used for inference.

In the \texttt{E4} framework, each DNN exit point corresponds to a classifier. 
For each exit point, the layer $l$ of each classifier $f_t$ within the given DNN layer indexed by frame $t$ is dynamically determined. 
Each classifier $f_t$ takes the accumulated features $z_t$ as input and predicts the final video label.
The training optimizes the parameters $\gamma$ of the classifier $f_t$ using the standard cross-entropy loss $\ell_{CE}$:
\begin{equation}
\mathcal{L}_{cls} = \frac{1}{T}\sum_{t=1}^T\ell_{CE}
\left(
f_t(z_t;\gamma), y 
\right),
\label{ext_cls}
\end{equation}
where $y$ is the label of the video frame.

In \texttt{E4}, each gate is implemented as a multi-layer perceptron (MLP) with low computational overhead. Specifically, the two-layer MLP processes the aggregated features $z_t$ and $z_{t-1}$ along with the normalized weights $\beta(\sigma_t)$, with 64 and 32 neurons per layer, respectively. 
We optimize the gate parameter $\mu$ using the binary cross-entropy (BCE) loss:
\begin{equation}
\mathcal{L}_{gate} = 
\frac{1}{T}\sum_{t=1}^T\ell_{BCE}
\left(
\rho((z_{t-1}, z_t,\beta(\sigma)_t);\mu),y
\right).
\label{eq4}
\end{equation}

We also use the standard cross-entropy (CE) loss to optimize the parameters $\sigma$ of the attention network $\beta$:
\begin{equation}
\mathcal{L}_{att}=\frac{1}{T}\sum_{t=1}^T\ell_{CE}(\beta(z_t;\sigma), y).
\label{eq5}
\end{equation}

The final loss to train the early-exit DNN model is:
\begin{equation}
\mathcal{L}=\mathcal{L}_{cls}+\mathcal{L}_{gate}+\mathcal{L}_{att}.
\label{eq6}
\end{equation}

\begin{algorithm}[t]
\caption{CDS-Based Frequency Scaling Algorithm}
\label{alg:alg1}
\textbf{Input}: The DNN exit points $e_t$, the CPU-GPU clock frequency range $[C_{min},C_{max}],[G_{min},G_{max}]$\\
\textbf{Output}: The optimal CPU-GPU clock frequency pair $(C_{f}^{*},G_{f}^{*})$ for each layer before $e_t$
\begin{algorithmic}[1] 
\STATE Initialize a dictionary $\mathcal{D} \leftarrow \{(C_{f_0},G_{f_0}); power\}$.
\FOR{round $r=1$ to $R$}
  \WHILE{$C_f \in [C_{min}, C_{max}]$ and $G_f \in [G_{min}, G_{max}]$}
    \FOR{$i=1$ to $e_t$}
      \STATE Sample $N$ candidates $(C_f, G_f)$.
      \FOR{$n$-th candidate}
        \STATE Profile the energy consumption $p$.
        \STATE $\mathcal{D} \leftarrow \{(C_f, G_f); power\}$
      \ENDFOR
      \STATE Update the CPU-GPU clock frequency to the one with the lowest energy consumption.
    \ENDFOR
  \ENDWHILE
\ENDFOR
\STATE Sort $\mathcal{D}$ by the profiled energy consumption.
\RETURN $(C_{f}^{*}, G_{f}^{*})$.
\end{algorithmic}
\end{algorithm}

\begin{table*}[t]
\caption{Configurations of edge devices used in our experiments.}
\label{tbl:device}
\centerline{
\footnotesize
\begin{tabular}{lccccccc} \hline
\textbf{Edge GPU}  & \textbf{Computing Power}  & \textbf{DRAM}  & \textbf{CPU} & \textbf{GPU}  & \textbf{Max Power} \\ \hline
Jetson Nano             & 0.47TFLOPS (FP16)    & 4GB             & 4$\times$Cortex-A57@1.4GHz     &  128$\times$Maxwell@0.9GHz   & 10W   \\
Jetson TX2              & 1.33TFLOPS (FP16)    & 8GB             & 6$\times$Cortex-A57@1.4GHz     &  256$\times$Pascal@1.3GHz      & 15W \\
Jetson Xavier NX        & 21TOPS (INT8)        & 8GB             & 6$\times$Carmel@1.4GHz         &  384$\times$Volta@1.1GHz       & 20W  \\
Jetson Orin Nano        & 40TOPS (INT8)        & 8GB             & 6$\times$Cortex-A78AE@1.5GHz   &  512$\times$Ampere@0.6GHz    & 15W   \\
Jetson AGX Orin         & 275TOPS (INT8)       & 64GB            & 12$\times$Cortex-A78AE@2.2GHz  &  2048$\times$Ampere@1.3GHz     & 60W  \\
\hline
\end{tabular}
}
\end{table*}

\subsection{Just-In-Time (JIT) Profiler}

After analyzing frame complexity and determining the DNN exit point, we implemented a \emph{Just-In-Time (JIT) Profiler} to efficiently identify the optimal power management configurations via coordinate descent search (CDS). CDS is a non-gradient optimization method focusing the search on one dimension at a time while keeping the values of other dimensions fixed. By alternating between dimensions, it converges to the optimal solution. This approach effectively transforms multi-variable optimization problems into single-variable ones, enhancing sampling efficiency. We also compare CDS performance with other methods like random search, evaluating both latency and energy consumption.


Algorithm~\ref{alg:alg1} outlines our CDS-based dynamic frequency scaling approach using dynamic voltage and frequency scaling (DVFS) technology. The algorithm takes DNN exit points $e_t$ and the CPU and GPU clock frequency ranges $[C_{min},C_{max}],[G_{min},G_{max}]$ as input. It initializes a dictionary $\mathcal{D}$ to track CPU and GPU clock frequencies along with the corresponding energy consumption $p$ for each round. For the searching of optimal settings, the CDS algorithm treats the layer before each DNN exit point as a separate coordinate. 
It then alternates the search for the optimal CPU-GPU clock frequencies for each layer, keeping other coordinates fixed at their previous optimal values. With each iteration, CPU-GPU clock frequencies are updated layer by layer. The process continues until all layers have been optimized, returning the best CPU-GPU clock frequency configuration. Compared to random search, CDS can find near-optimal solutions more quickly, as demonstrated in our performance evaluation experiments. 

In summary, the CDS-based JIT Profiler effectively finds the optimal low-power CPU and GPU clock frequencies for each layer before the DNN exit points, ensuring low-latency. 


\begin{figure*}[t]
\centerline{
\subfigure[EfficientNet-B0 on ActivityNet-v1.3 dataset]{
\begin{minipage}[b]{\linewidth}
\includegraphics[width=1\linewidth]{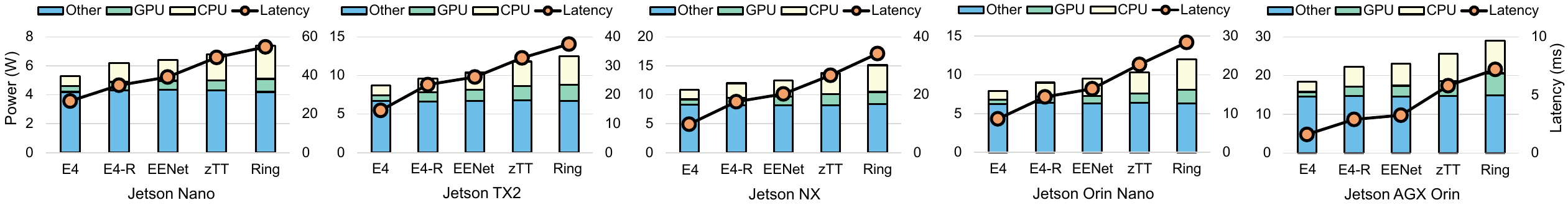}
\end{minipage}}
}
\centerline{
\subfigure[MobileNet-v2 on the ActivityNet-v1.3 dataset]{
\begin{minipage}[b]{\linewidth}
\includegraphics[width=1\linewidth]{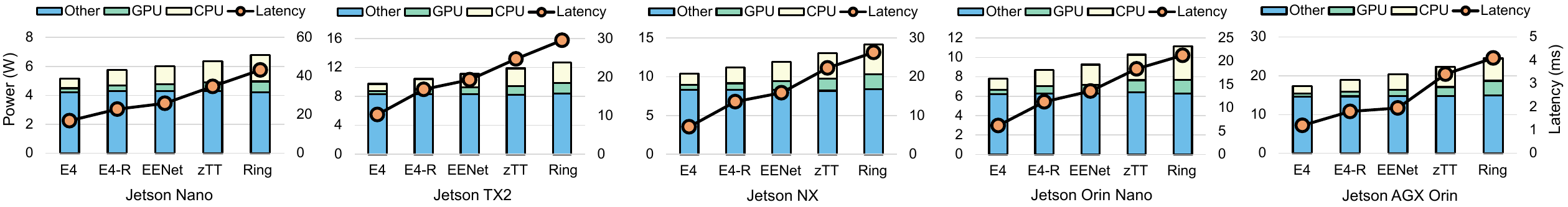}
\end{minipage}}
}
\centerline{
\subfigure[EfficientNet-B0 on the Mini-Kinetics dataset]{
\begin{minipage}[b]{\linewidth}
\includegraphics[width=1\linewidth]{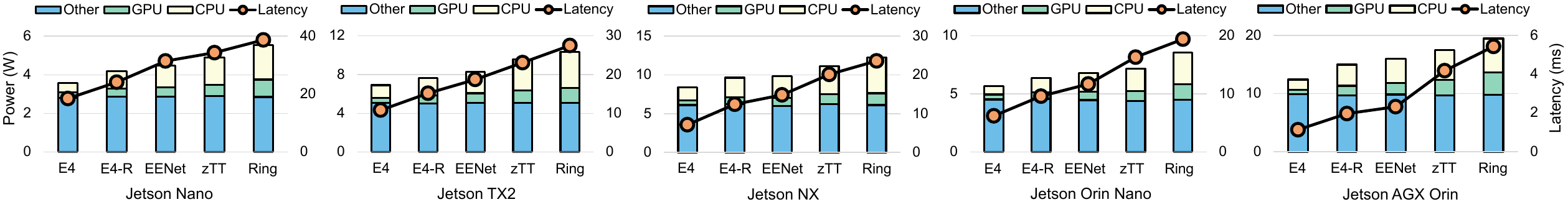}
\end{minipage}}
}
\centerline{
\subfigure[MobileNet-v2 on the Mini-Kinetics dataset]{
\begin{minipage}[b]{\linewidth}
\includegraphics[width=1\linewidth]{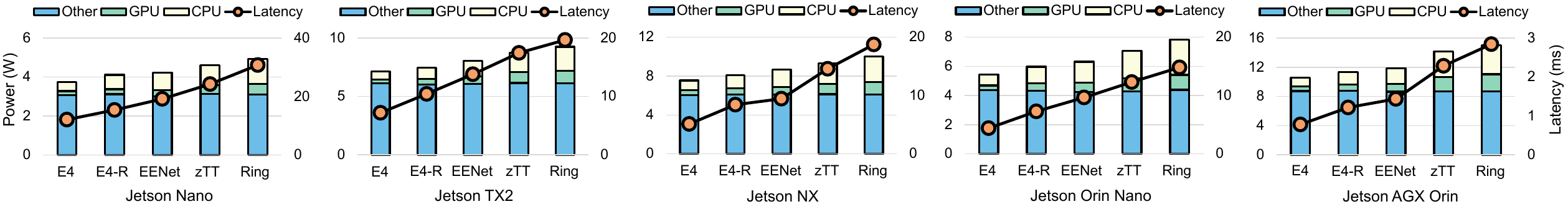}
\end{minipage}}
}
\caption{Comparison of energy consumption and inference latency for \texttt{E4} {\em vs.} baseline approaches (EENet, zTT, and Ring).}
\label{latency_energy}
\end{figure*}

\subsection{DVFS Governor Settings}

We next describe how we customize the DVFS governor specifically for edge video analytics, adapting it to both video frames and DNN models. 
Specifically, we use the Nvidia \texttt{nvpmodel}~\cite{nvpmodel}, a performance governor for the Jetson series edge devices, which allows users to dynamically adjust the CPU and GPU clock frequencies.
For example, the Jetson AGX Orin allows its CPU and GPU clock frequencies to scale between 0.1 and 2.2GHz, and 0.1 to 1.13GHz, respectively. 
Based on the identified DNN exit points, the DVFS governor dynamically adjusts the CPU and GPU clock frequencies according to the optimal power setting provided by the JIT Profiler.
Since DVFS governor operates with millisecond-level latency, the primary overhead is attributed to the JIT Profiler, which we will evaluate in detail later.

\section{Implementation and Evaluation}

\subsection{Implementation of E4}

\textbf{\texttt{E4} prototype:}
We implemented \texttt{E4} using Python 3.6 across five heterogeneous Nvidia edge devices. 
The specific configurations of edge devices are summarized in Table~\ref{tbl:device}. 
The minimum clock frequency for all edge devices is set to 0.1GHz to ensure basic system operation. 
We utilize \\texttt{jetson-stats}~\cite{jetson-stats} to measure the energy consumption of edge devices during DNN inference.

\medskip
\noindent\textbf{Baselines:}
We compare \texttt{E4} with the following alternatives. 

\noindent
$\bullet$
\textbf{EENet}~\cite{li2023eenet} is a state-of-the-art energy efficient inference method incorporating early-exit and DVFS. It provides frequency and voltage calibration advice over short timescales by predicting where the DNN model will exit based on inference workloads and timing constraints.

\noindent
$\bullet$
\textbf{zTT}~\cite{kim2021ztt} is a recent workload-aware DVFS governor that utilizes deep reinforcement learning (DRL) to adopt to ambient temperature, aiming to optimize CPU and GPU clock frequencies for energy savings.

\noindent
$\bullet$
\textbf{Ring-DVFS}~\cite{yeganeh2020ring} is a DRL enhanced DVFS governor designed for multi-core embedded systems, aimed at reducing energy consumption.
This approach is applied to edge devices with heterogeneous CPU-GPU  processors using Nvidia \texttt{nvpmodel}~\cite{nvpmodel} for a fair comparison.

\noindent
$\bullet$
\textbf{E4-R} is a reduced baseline of our proposed \texttt{E4} with CDS replaced with random search. 
Random search samples CPU-GPU clock frequencies and power management configurations randomly, and profiles their energy consumption as the cost. A cache is used to record all schedules and their associated costs. After a specified number of search rounds, random search returns the schedule with the lowest found energy consumption.

\subsection{Experimental Setup}
\label{setup}

\textbf{DNN models:}
We utilize EfficientNet-B0~\cite{tan2019efficientnet} and MobileNet-v2~\cite{sandler2018mobilenetv2} for edge video analytics.
Both DNN models are pretrained on the ImageNet dataset, with five DNN exit points configured for each model. 
We remove the last classification layer from the backbone and replace it with a fully-connected layer with 1024 neurons. 
The early-exit DNN models are trained offline using four Nvidia 3080 GPUs with a mini-batch size of 512. 
Training is conducted with the Adam optimizer and a learning rate of $10^{-4}$. 

\medskip
\noindent\textbf{Datasets:}
We conduct experiments on two large-scale datasets: ActivityNet-v1.3~\cite{caba2015activitynet} and 
Mini-Kinetics~\cite{kay2017kinetics}.
ActivityNet-v1.3 is a long-range action recognition dataset comprising 20,000 videos across 200 classes (10,024 videos for training, 4,926 for validation, and 5,044 for testing).
Each video has an average duration of 167 seconds and 1.5 labels. 
Mini-Kinetics, provided by~\cite{meng2020ar}, is a short-range action recognition dataset featuring 200 classes from the Kinetics dataset~\cite{caba2015activitynet}, with 121,215 training and 9,867 test videos, each averaging 10 seconds in duration. 
We use \emph{top-1 accuracy} to evaluate multi-label classification performance in ActivityNet, and \emph{mean average precision (mAP)} for multi-class classification on Mini-Kinetics.

\subsection{Performance Evaluation}
\label{Evaluation}

\textbf{Comparison of energy and latency}: 
We compare the inference performance of \texttt{E4} with other methods across two DNN models and two datasets, using the five heterogeneous edge devices listed in Table~\ref{tbl:device}.

Fig.~\ref{latency_energy}(a,c) shows that for video analysis with EfficientNet-B0~\cite{tan2019efficientnet}, energy consumption is primarily driven by non-video analytics system operation (denoted as ``Other''), followed by CPU and GPU usage for data preprocessing, transmission, and parallel computing.
\texttt{E4} improves energy efficiency, achieving 20\% to 37\% energy saving and a $1.3\times$ to $2.2\times$ speedup. The performance improvement of \texttt{E4} is attributed to DNN's early-exit mechanism, which significantly reduces computation cost and energy consumption by partial DNN inference. 
In comparison, zTT~\cite{kim2021ztt} and Ring-DVFS~\cite{yeganeh2020ring} focus solely on optimizing the clock frequency of individual heterogeneous processors. 
Even compared to EENet~\cite{li2023eenet}, a state-of-the-art energy-efficient inference framework that coordinates early exit and DVFS, \texttt{E4} has lower inference latency and higher energy saving, thanks to an attention-based cascade that cautiously guides early exit and DVFS governor by analyzing the complexity of video frames.

In Fig.~\ref{latency_energy}(b) and Fig.~\ref{latency_energy}(d), we further examine the video analytics task using MobileNet-v2~\cite{sandler2018mobilenetv2}. 
It can be observed that \texttt{E4} reduces 18\%$\sim$30\% energy consumption and achieves $1.4\times$$\sim$$1.9\times$ inference speedup, respectively. 
Since the computation complexity of MobileNet-v2~\cite{sandler2018mobilenetv2} is lower than EfficientNet-B0~\cite{tan2019efficientnet}, the search space of our CDS-based \emph{JIT Profile} is limited, which causes the energy-efficient improvement to be slightly lower than EfficientNet-B0~\cite{tan2019efficientnet}.

Moreover, \texttt{E4} has better performance than \texttt{E4-R} for five heterogeneous edge devices. 
The results reveal that although the random search algorithm is simple, it could significantly reduce the runtime energy consumption and latency, highlighting the advantages of our framework design.
Interestingly, we find that compared with edge devices with low computing power, \texttt{E4} brings more significant performance improvement to edge devices with high computing power. 
For instance, the performance improvement of Jetson AGX Orin, which has the highest computing power, is 37\% higher on average than that of Jetson Nano with the lowest computing power.
The results are attributed to the fact that high computing power means a larger frequency range, so that \texttt{E4} has a larger optimization space.

\begin{figure}[t]
\centerline{
\includegraphics[width=0.5\linewidth]{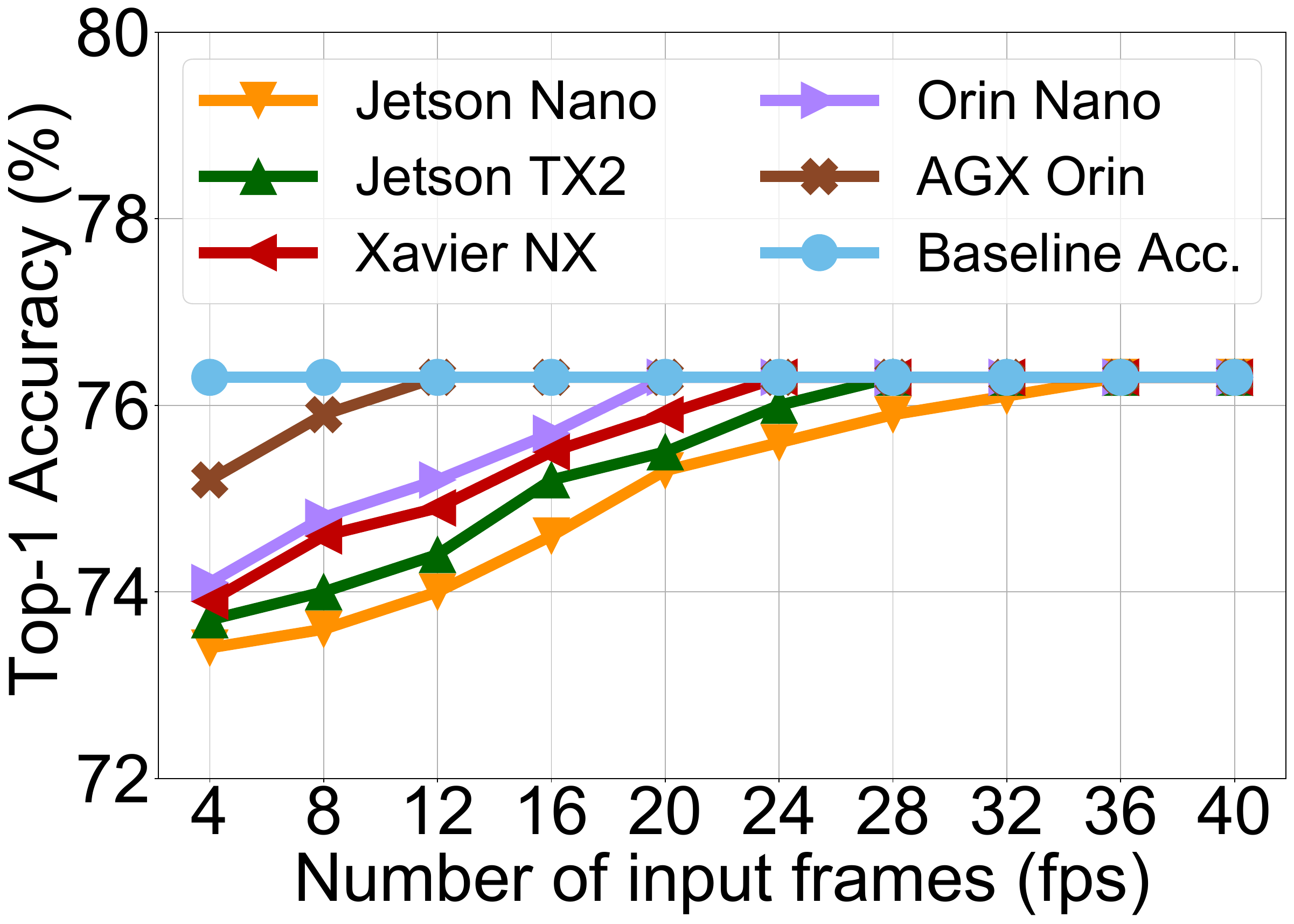}
\includegraphics[width=0.5\linewidth]{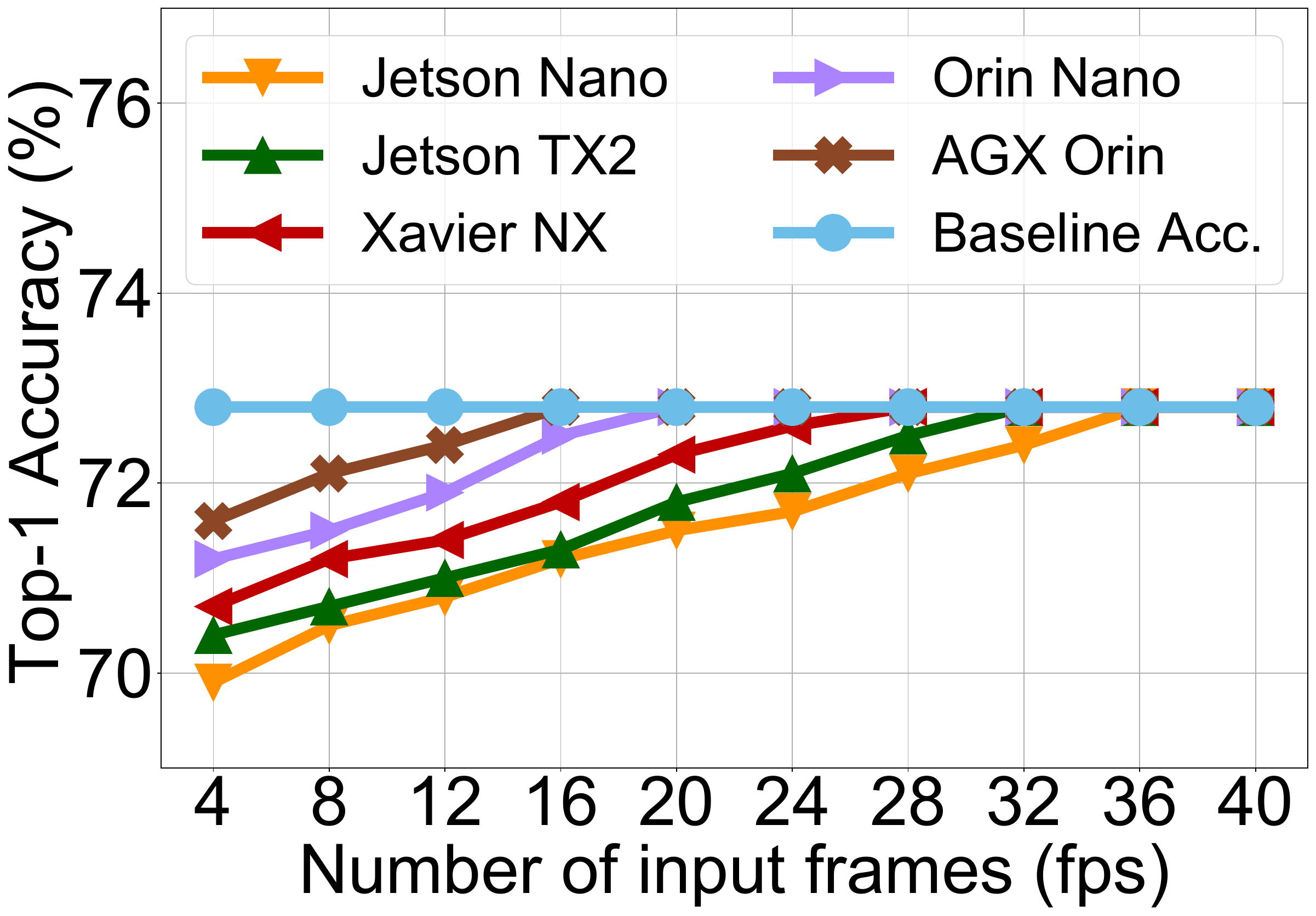}
}
\centerline{
\footnotesize{
(a) EfficientNet-B0 \hspace{18mm} (b) MobileNet-v2
}}
\centerline{
\includegraphics[width=0.5\linewidth]{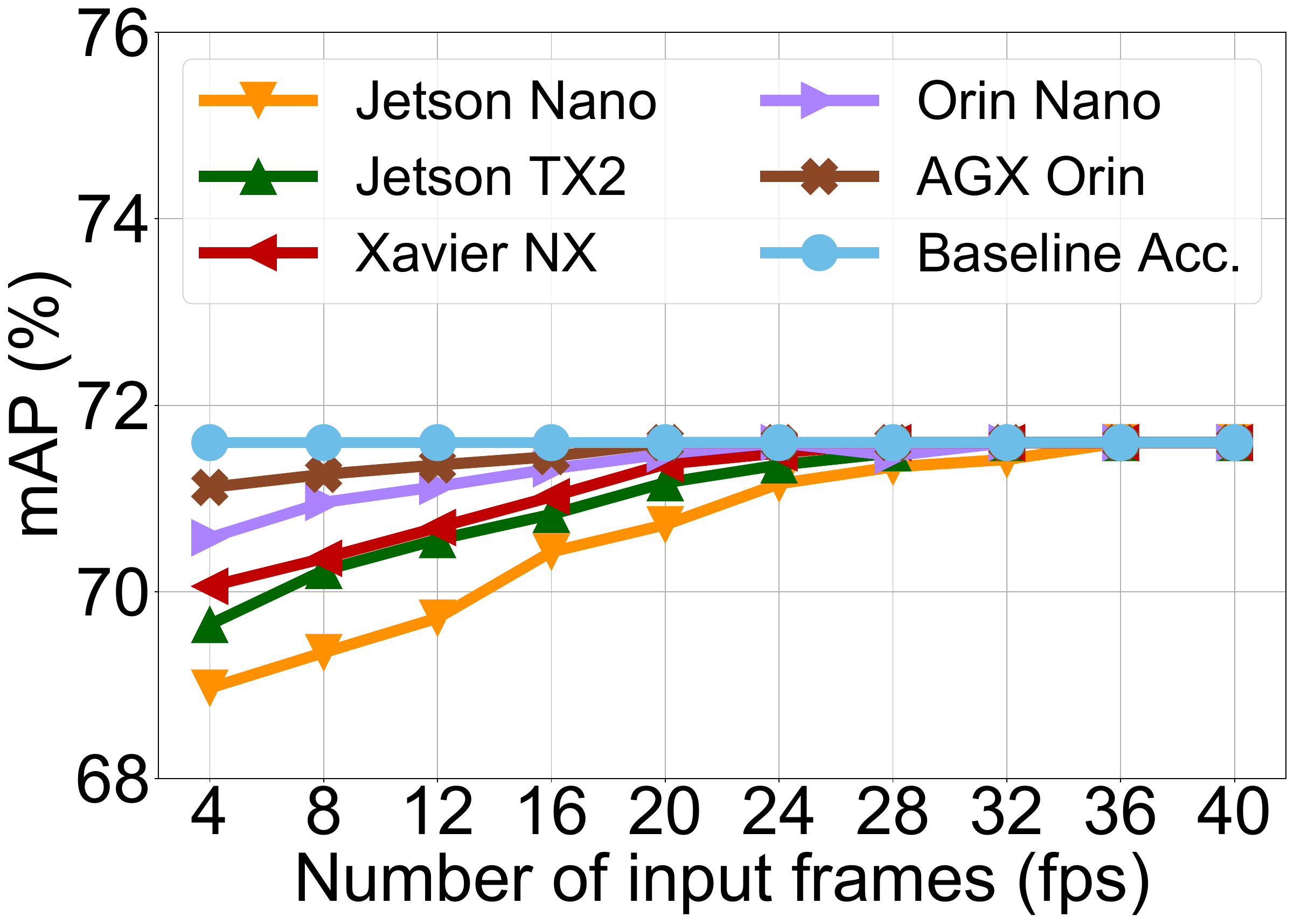}
\includegraphics[width=0.5\linewidth]{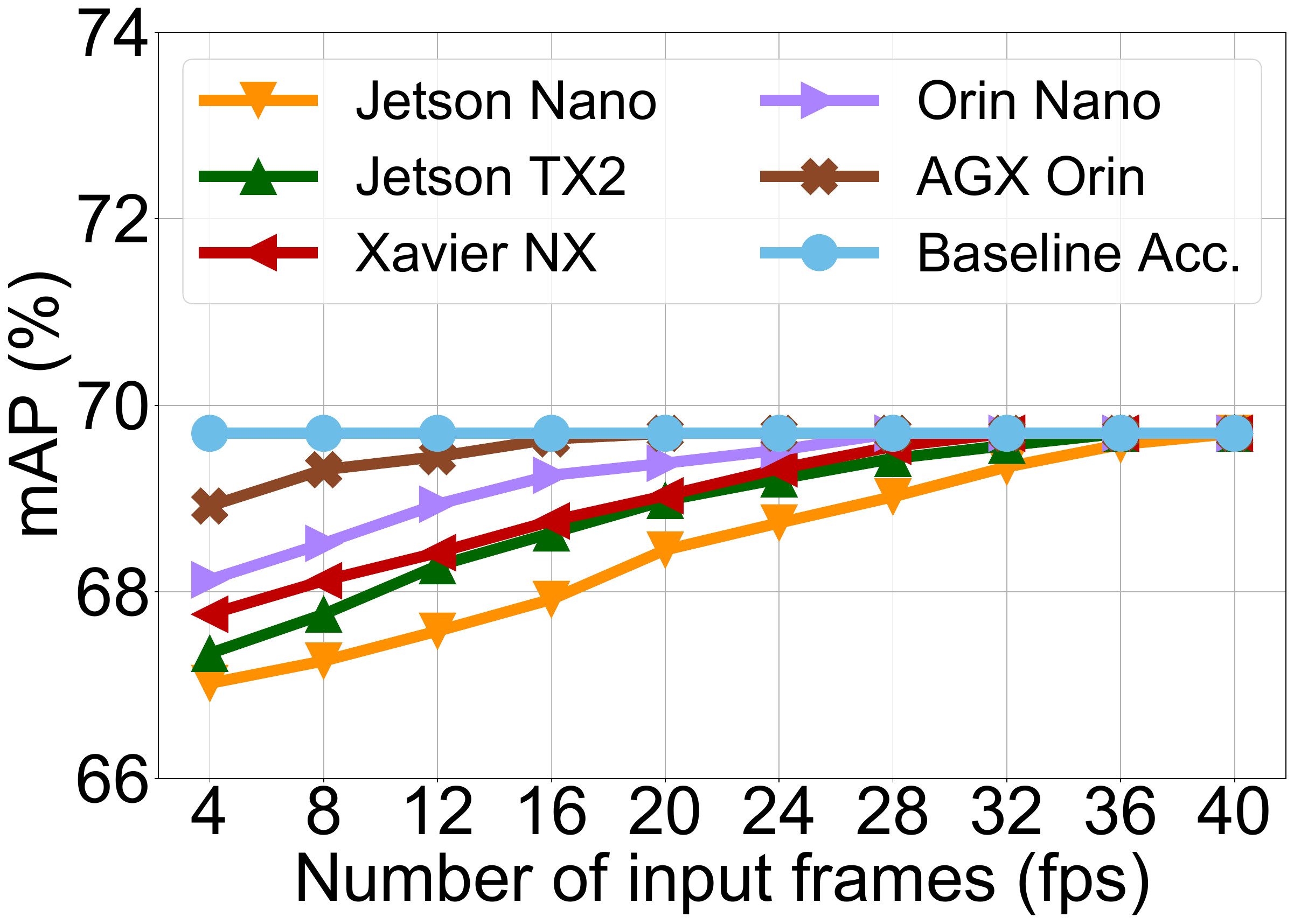}
}
\centerline{
\footnotesize{
(c) EfficientNet-B0 \hspace{18mm} (d) MobileNet-v2
}}
\caption{Effect of input frame rate {\em vs.} inference accuracy on (a,b) ActivityNet-v1.3 and (c,d) Mini-Kinetics datasets.}
\label{accuracy}
\end{figure}

\medskip
\noindent
\textbf{Impact of number of input frames on accuracy:}
The accuracy of early-exit in \texttt{E4} is closely related to the number of input frames. 
Intuitively, the higher the number of input frames, the richer the information of the frames extracted by the aggregated feature pool, which means higher accuracy, but higher computational overhead.
We evaluate the accuracy of \texttt{E4} by scaling the number of input frames. 
As shown in Fig.~\ref{accuracy}, we report the accuracy of \texttt{E4} on ActivityNet and Mini-Kinetics dataset with various frame rates, respectively.
In particular, edge devices with high computing power can process more video frames within the same time window, which means better adaptability to frame rates ($e.g.$, the accuracy loss of Jetson AGX Orin is within 1\%).
For instance, \texttt{E4} leverages more video frames for complex objects, and conversely uses fewer frames to inference simple objects. 
It can also be observed that the accuracy of \texttt{E4} improves with the number of input frames , but up tp a certain limit. 
On the one hand, complex objects may require more frames to be recognized. 
On the other hand, the reason why an excessively high number of frames cannot further improve the accuracy  is attributed to the inability of convolutional neural networks in leveraging temporal information. 
Therefore, we set $T = 20$ in our experiments to keep the balance between accuracy and computational overhead.

\medskip
\noindent
\textbf{Ablation study:}
We inspect the performance improvement of different components in \texttt{E4}, including the CDS-based DVFS governor and the attention-based early-exit framework. 
For all ablations, we follow the same training procedure explained in Section Experimental Setup.
Jetson AGX Orin is uesd for DNN inference on Activitynet-v1.3 and Mini-Kinetics dataset, respectively. 
Table~\ref{ablation} reports the energy consumption and inference latency for all combinations. 
We first evaluate the behavior of DVFS governor. 
Despite increasing 12\% inference latency on average, DVFS governor reduces 26\% energy consumption, which is desirable. 
We then inspect the impact of attention-based DNN's early-exit mechanism. 
It can be observed that inference latency and energy consumption are significantly reduced by 31\% and 28\% respectively, which is attributed to partial DNN inference, achieving energy saving while reducing computation cost. 
Compared with the original DNN inference without DVFS and early-exit, the performance improvement of \texttt{E4} is the most significant, reducing the energy consumption and inference latency by up to 46\% and 73\%, respectively.
It reveals the effective complementarity of DVFS governor and DNN's early-exit mechanism in co-optimizing energy consumption and inference latency.

\begin{table}[t]
\caption{Ablation studies of DVFS and early-exit on Nvidia Jetson AGX Orin.}
\label{ablation}
\centerline{
\footnotesize
\begin{tabular}{c|cc|cc}
\hline
\textbf{DNN} & \textbf{DVFS} & \textbf{Early-Exit} & \textbf{Latency (ms)} & \textbf{Power (W)} \\
\hline
\multirow{4}{*}{\makecell[c]{Eff-B0 \\ on Act}} & \textcolor{purple}{\ding{55}} & \textcolor{purple}{\ding{55}} & 6.3 & 34.6 \\
& \textcolor{teal}{\checkmark} & \textcolor{purple}{\ding{55}} & 7.1 & 23.1 \\
& \textcolor{purple}{\ding{55}} & \textcolor{teal}{\checkmark} & 4.6 & 24.5 \\
& \textcolor{teal}{\checkmark} & \textcolor{teal}{\checkmark} & \textbf{1.6} & \textbf{18.4} \\
\hline
\multirow{4}{*}{\makecell[c]{Mob-v2 \\ on Act}} & \textcolor{purple}{\ding{55}} & \textcolor{purple}{\ding{55}} & 4.1 & 27.3 \\
& \textcolor{teal}{\checkmark} & \textcolor{purple}{\ding{55}} & 4.9 & 21.6 \\
& \textcolor{purple}{\ding{55}} & \textcolor{teal}{\checkmark} & 3.4 & 23.1 \\
& \textcolor{teal}{\checkmark} & \textcolor{teal}{\checkmark} & \textbf{1.2} & \textbf{17.3} \\
\hline
\multirow{4}{*}{\makecell[c]{Eff-B0 \\ on Mini}} & \textcolor{purple}{\ding{55}} & \textcolor{purple}{\ding{55}} & 3.9 & 23.7 \\
& \textcolor{teal}{\checkmark} & \textcolor{purple}{\ding{55}} & 4.3 & 18.2 \\
& \textcolor{purple}{\ding{55}} & \textcolor{teal}{\checkmark} & 2.4 & 16.7 \\
& \textcolor{teal}{\checkmark} & \textcolor{teal}{\checkmark} & \textbf{1.1} & \textbf{12.4} \\
\hline
\multirow{4}{*}{\makecell[c]{Mob-v2 \\ on Mini}} & \textcolor{purple}{\ding{55}} & \textcolor{purple}{\ding{55}} & 3.2 & 22.7 \\
& \textcolor{teal}{\checkmark} & \textcolor{purple}{\ding{55}} & 3.6 & 16.9 \\
& \textcolor{purple}{\ding{55}} & \textcolor{teal}{\checkmark} & 1.7 & 14.2 \\
& \textcolor{teal}{\checkmark} & \textcolor{teal}{\checkmark} & \textbf{0.8} & \textbf{10.6} \\
\hline
\end{tabular}
}
\end{table}

\begin{figure}[t]
\centerline{
\includegraphics[width=\linewidth]{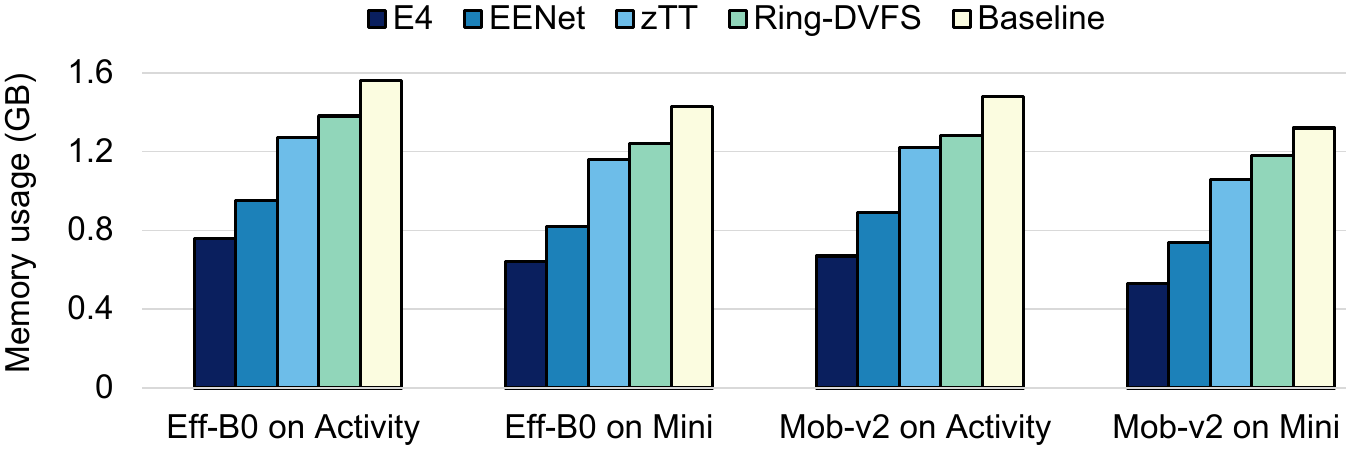}
}
\caption{Comparison of memory usage between \texttt{E4} and other approaches using two DNN models for edge video analysis on the ActivityNet-v1.3 and Mini-Kinetics datasets.}
\label{memory}
\end{figure}

\textbf{Memory usage:}
As shown in Fig.~\ref{memory}, we report the memory usage of \texttt{E4} compared to other approaches. 
Clearly, the memory usage of \texttt{E4} consistently outperforms other approaches. 
Compared to the baseline (without early exit and DVFS), the average memory usage is reduced by up to 55\%, which is promising for resource-constrained (especially memory) edge devices, further facilitating the deployment of video analytics to edge devices or even microprocessor units (MCU) with lower resource and power consumption, thereby reducing costs. 
Furthermore, \texttt{E4} reduces memory usage by 45\%$\sim$49\% compared to \emph{zTT}~\cite{kim2021ztt} and \emph{Ring-DVFS}~\cite{yeganeh2020ring} without early exit, which is mainly attributed to the memory saving from early exit. 
In comparison, the performance improvement of DVFS technology for energy saving is limited, 
saving only 6\%$\sim$10\% in memory usage compared to baseline.
Although \emph{EENet}~\cite{li2023eenet} coordinates early exit and DVFS, it does not focus on the complexity of video frames, thus increasing memory usage by 23\% compared to \texttt{E4}. 
Overall, early exit dominates memory usage, while the DVFS technology can further reduce memory usage by fine-grained tuning of the CPU and GPU clock frequencies.

\section{Conclusions}
\label{Conclusion}

In this paper, we propose \texttt{E4}, an energy-efficient DNN inference framework for edge video analytics. 
\texttt{E4} introduces two design knobs to enable energy-efficient DNN inference: the early-exit mechanism of DNN models and a Just-in-Time profiler to obtain optimal CPU and GPU clock frequency configurations. 
Comprehensive experiments with prototype implementations on heterogeneous edge devices show that \texttt{E4} outperforms state-of-the-art early-exit approaches in terms of energy consumption and inference latency without sacrificing accuracy significantly. 
We earnestly hope that \texttt{E4} will inspire the community to consider energy as a first-class resource in DNN optimization in edge video analytics.

\section{Acknowledgments}
\label{Acknowledgments}
This work was supposed by the National Natural Science Foundation of China under Grant No. 62350710797 and the National Key R\&D Program of China under Grant No. 2022YFF0503900.

\bibliography{aaai25}

\end{document}